\documentclass{article}
\usepackage[utf8]{inputenc}
\usepackage{amsmath, amssymb, amsthm}
\usepackage{algorithm}
\usepackage{algpseudocode}
\usepackage{geometry}
\usepackage{graphicx}
\usepackage{natbib}
\usepackage{booktabs}
\usepackage{booktabs}
\usepackage{array}     
\usepackage{xcolor}
\usepackage{subcaption}
\usepackage{url}
\usepackage{tikz}
\usepackage{pgfplots}
\usepackage{hyperref}

\newcolumntype{P}[1]{>{\raggedright\arraybackslash}p{#1}}

\title{Training data membership inference via Gaussian process meta-modeling: a post-hoc analysis approach}
\author{
Yongchao Huang\footnote{\texttt{yongchao.huang@abdn.ac.uk}}
\and
Pengfei Zhang\footnote{\texttt{pf.zhang@binance.com}}
\and
Shahzad Mumtaz\footnote{\texttt{shahzad.mumtaz@abdn.ac.uk}}
}

\date{May 2025}

\begin{document}

\maketitle

\begin{abstract}
Membership inference attacks (MIAs) test whether a data point was part of a model's training set, posing serious privacy risks. Existing methods often depend on shadow models or heavy query access, which limits their practicality. We propose GP-MIA, an efficient and interpretable approach based on Gaussian process (GP) meta-modeling. Using post-hoc metrics such as accuracy, entropy, dataset statistics, and optional sensitivity features (e.g. gradients, NTK measures) from a single trained model, GP-MIA trains a GP classifier to distinguish members from non-members while providing calibrated uncertainty estimates. Experiments on synthetic data, real-world fraud detection data, CIFAR-10, and WikiText-2 show that GP-MIA achieves high accuracy and generalizability, offering a practical alternative to existing MIAs.
\end{abstract}

\section{Introduction}

Machine learning models, particularly deep neural networks, are vulnerable to privacy risks in the form of membership inference attacks (MIAs). 
An adversary carrying out an MIA attempts to determine whether a specific data point was included in a model’s training dataset \cite{Shokri2017MIA, Ye2022EnhancedMIA, Carlini2022FirstPrinciples}. 
Such attacks exploit the systematic behavioral differences that models exhibit on training versus unseen data, for example, higher confidence, lower loss, or sharper decision boundaries on members. 
These vulnerabilities raise serious concerns for sensitive domains such as healthcare, finance, or security, where disclosing whether an individual’s record was used for training could already constitute a privacy breach. 

Formally, let $f_{\theta^*}:\mathbb{R}^d \to \mathbb{R}^m$ denote a model with parameters $\theta^*$, trained on a dataset 
$X = \{(x_i, y_i)\}_{i=1}^n$. 
Given a new test pair $(x, y)$, the membership inference task is to decide whether this pair was part of $X$. 
That is, the goal is to design an inference rule
\[
M(f_{\theta^*},x,y) \in \{0,1\},
\]
which outputs $1$ if $(x,y)$ is inferred to belong to the training set and $0$ otherwise. 
The challenge lies in exploiting subtle statistical signals that separate members from non-members, while avoiding strong assumptions such as access to the model’s parameters, gradients, or training pipeline.

State-of-the-art MIA methods vary in how they access and exploit the target model. 
Shadow model-based approaches, introduced by Shokri et al. \cite{Shokri2017MIA}, train multiple auxiliary models to mimic the target’s behavior and then build a binary classifier to distinguish membership. 
Other works, such as the likelihood ratio attack (LiRA) of Carlini et al. \cite{Carlini2022FirstPrinciples}, adopt hypothesis testing frameworks based on model output distributions. 
Although effective, these techniques typically require heavy computational resources, carefully curated auxiliary data, or multiple queries to the target model. 
Such requirements limit their practicality in realistic, possibly large, black-box settings where only model outputs are observable.

\paragraph{Our contribution.}
We propose a new post-hoc MIA method, GP-MIA, that employs Gaussian processes (GPs) as meta-models to detect membership from statistical features of datasets passed through a trained model. 
Specifically, we pass the test data point to the target model, and extract diagnostic metrics such as accuracy, prediction entropy, and input statistics (e.g., feature means and variances). 
Also, lightweight perturbation-based features can be included to improve inference accuracy. 
These metrics form feature vectors that are used to train a GP classifier to distinguish member from non-member datasets. 
Unlike shadow-based or gradient-based methods, GP-MIA requires only post-hoc access to model predictions, avoids retraining or internal access, and naturally provides calibrated uncertainty estimates, enhancing both efficiency and interpretability. 
This makes GP-MIA a practical tool for privacy auditing in real-world deployment scenarios.

\section{Related Work}

Membership inference attacks have been studied extensively since the advent of machine learning and big data, with early works in the 1990s and 2000s exploring privacy risks in statistical databases. Foundational ideas by Dalenius  \cite{Dalenius1977towards} laid the groundwork for understanding disclosure risks in aggregated data. Agrawal and Srikant  \cite{Agrawal2000privacy} proposed one of the first privacy-preserving data mining frameworks, and Homer et al. \cite{Homer2008Resolving} demonstrated how individual genomic membership could be inferred from statistical aggregates. These studies established early awareness of the privacy vulnerabilities that later evolved into attacks on modern machine learning models. In the era of machine learning, Shokri et al. \cite{Shokri2017MIA} introduced MIAs by training multiple shadow models to mimic a target model's behavior, using their outputs to train a binary classifier for membership inference; this method achieves high accuracy but is computationally expensive due to the need for numerous shadow models, limiting scalability in resource-constrained settings. Ye et al. \cite{Ye2022EnhancedMIA} enhanced this framework by incorporating features like loss distributions and confidence scores, improving attack success rates for deep neural networks in both black-box and white-box scenarios; however, it still relies on shadow models and requires carefully curated auxiliary datasets, reducing efficiency and generalizability. Carlini et al. \cite{Carlini2022FirstPrinciples} developed the likelihood ratio attack (LiRA), which employs a hypothesis-testing framework to compare the likelihood of a data point being a member versus non-member based on model outputs, excelling at low false-positive rates; yet, LiRA demands multiple model queries and significant computational resources for calibration, making it less practical for black-box settings. Hu et al. \cite{Hu2022MIA} provided a comprehensive survey, categorizing MIAs by strategies (e.g. shadow models, threshold-based, metric-based) and access levels, emphasizing trade-offs between accuracy, computational cost, and model access, and calling for efficient, interpretable methods. These works address the challenges of computational overhead and model access in MIAs.

Our proposed Gaussian process-based MIA departs from these approaches by eliminating the reliance on shadow models or extensive query access. Instead, it builds on post-hoc metrics derived from a single trained model, such as accuracy, entropy, and loss, complemented by lightweight dataset statistics and perturbation magnitudes. This can be further extended to include sensitivity-based features (e.g., gradient norms and NTK statistics), providing a richer signal of membership. The GP classifier leverages these metrics to distinguish members from non-members while also quantifying uncertainty, yielding a method that is efficient, interpretable, and well-suited to practical settings where adversaries have only limited access to model outputs or gradients.

\section{Preliminaries on Gaussian processes}

A Gaussian process (GP \cite{Rasmussen2004GP,Rasmussen2006GPbook}) is a non-parametric probabilistic model defined by a mean function $m(\mathbf{x})$ and a covariance (kernel) function $k(\mathbf{x}, \mathbf{x}')$:

\[
f(\mathbf{x}) \sim \mathcal{GP}(m(\mathbf{x}), k(\mathbf{x}, \mathbf{x}')).
\]

It can be used for both regression (e.g., time series forecasting) and classification. 
For regression, the GP directly models the distribution of outputs with Gaussian likelihood. 
For binary classification, however, a GP is combined with a non-Gaussian likelihood (e.g., Bernoulli) to model class probabilities. 
The posterior predictive distribution for a new input $\mathbf{x}_*$ is:

\[
p(y_* \mid \mathbf{x}_*, \mathcal{D}) 
= \int \sigma(f(\mathbf{x}_*)) \, p(f(\mathbf{x}_*) \mid \mathbf{x}_*, \mathcal{D}) \, df(\mathbf{x}_*),
\]

where $\sigma(\cdot)$ is a sigmoid (or probit) link function and $\mathcal{D}$ denotes the training data. 
Because this integral is analytically intractable, approximate inference methods such as the Laplace approximation, expectation propagation, or variational inference are typically employed. 

A common kernel choice is the radial basis function (RBF) kernel:

\[
k(\mathbf{x}, \mathbf{x}') = \sigma^2 \exp\!\left(-\tfrac{1}{2\ell^2}\|\mathbf{x} - \mathbf{x}'\|^2\right),
\]

with variance $\sigma^2$ and lengthscale $\ell$ controlling the smoothness of the latent function. 
The RBF kernel can be augmented with a white noise kernel to account for observation noise.

\paragraph{GP classifier.}  
In practice, a GP classifier maps input features into a latent GP function, then squashes this latent function through a sigmoid to obtain class probabilities. 
Given feature vectors $\{\mathbf{x}_i\}$ and binary labels $\{y_i\}$, the GP defines a joint prior distribution over latent values $\mathbf{f} = [f(\mathbf{x}_1), \dots, f(\mathbf{x}_n)]^\top$. 
Training consists of optimizing the kernel hyperparameters by maximizing the marginal likelihood (or its variational lower bound). 
At prediction time, the GP outputs calibrated class probabilities that naturally reflect predictive uncertainty. 

GPs are well-suited for membership inference due to their ability to model uncertainty and capture subtle separability in high-dimensional feature spaces. 
In our approach, the GP classifier takes as input scalar metrics extracted from a trained neural network and outputs membership probabilities, to effectively distinguish between member and non-member data samples.

\section{GP-MIA: Methodology}

We present GP-MIA, a post-hoc framework for detecting whether individual data points or datasets contributed to the training of a supervised model. The method is model-agnostic and requires no access to the training process, only a trained target model $f_{\theta^*}$ that has been brought to convergence. This frozen model serves as an oracle, from which we extract diagnostic metrics characterizing its behavior on candidate inputs. These metrics are then used to train a Gaussian Process (GP) classifier that distinguishes between member and non-member data. At test time, a new input is passed through the target model, its metrics are computed in a single forward (and optionally backward) scan, and the GP returns a calibrated membership probability with uncertainty estimates.  

\paragraph{Feature construction.}  
The first step in GP-MIA is to compute a set of diagnostic features that summarize how a trained model $f_{\theta^*}$ behaves on a candidate dataset or input $x$. These features are designed to capture differences in the statistical profile of member and non-member data. In its simplest configuration, as used in the synthetic and CIFAR-10 experiments, GP-MIA extracts performance-based metrics such as classification accuracy (or mean squared error in regression), model confidence measures such as the average entropy of predicted probabilities, and scalar summary statistics of the input distribution (mean and variance). We also consider a lightweight sensitivity proxy based on perturbation magnitude: the $\ell_2$ distance between the original model weights and those obtained after a few fine-tuning steps on the candidate dataset. Together, these common features provide a compact, model-agnostic signature that does not require access to gradients or internal activations, and already yield strong separability between members and non-members (as seen later).

Beyond these basic features, we extend GP-MIA by incorporating gradient-based and kernel-based diagnostics. For a trained neural network $f_{\theta^*}:\mathbb{R}^d \to \mathbb{R}^m$, the \textit{parameter-Jacobian} 
\[
g_\theta(x) = \nabla_\theta f_{\theta^*}(x) \in \mathbb{R}^{m \times p}
\]
captures the sensitivity of the model output with respect to the parameters. With labels $y$, one may also consider the loss gradient 
\[
g_\ell(x,y) = \nabla_\theta \ell(f_{\theta^*}(x),y) \in \mathbb{R}^p,
\]
while the \textit{input-Jacobian}
\[
J_x(x) = \frac{\partial f_{\theta^*}(x)}{\partial x} \in \mathbb{R}^{m \times d}
\]
quantifies local smoothness in input space. These yield a feature vector
\[
\phi_{\mathrm{grad}}(x) = \big[ \|g_\theta(x)\|_{\mathrm{F}}, \; \|J_x(x)\|_{\mathrm{F}}, \; \ell(f_{\theta^*}(x),y), \; \|g_\ell(x,y)\|_2 \big].
\]

We further exploit the neural tangent kernel (NTK) geometry, where the kernel between two points is
\[
k_{\theta^*}(x,x') = g_\theta(x) g_\theta(x')^\top,
\]
from which leverage scores
\[
\tau_\lambda(x) = k_x^\top (K+\lambda I)^{-1} k_x
\]
and projection statistics $h_\lambda(x)$ can be computed. These yield the vector
\[
\phi_{\mathrm{ntk}}(x) = \big[ \tau_\lambda(x), \; \|h_\lambda(x)\|_2, \; \max_i |h_\lambda(x)_i|, \; s_{\max}(x), \; \bar{s}(x) \big].
\]

Finally, a unified feature representation is obtained by concatenating basic features, gradient features, and NTK features:
\[
\phi(x) = \big[ \phi_{\mathrm{common}}(x), \; \phi_{\mathrm{grad}}(x), \; \phi_{\mathrm{ntk}}(x) \big].
\]
In practice, not all feature families are required. Our synthetic and CIFAR-10 experiments used only $\phi_{\mathrm{common}}$, while the language model experiment additionally incorporated $\phi_{\mathrm{grad}}$ to capture gradient-based sensitivity.

\paragraph{GP classification.}  
Once features are extracted, they are paired with binary membership labels ($1$ for member, $0$ for non-member) and used to train a Gaussian Process classifier with an RBF + white noise kernel. For binary classification, the GP posterior is combined with a Bernoulli likelihood, yielding predictive probabilities for membership:
\[
p(y_* = 1 \mid x_*, \mathcal{D}) = \int \sigma(f(x_*)) \, p(f(x_*) \mid x_*, \mathcal{D}) \, df(x_*),
\]
where $\sigma(\cdot)$ is the sigmoid link function. This design allows the GP to generalize from observed behavioral patterns to new candidate inputs, while expressing uncertainty when evidence is ambiguous. The GP-MIA procedure is summarized \footnote{A more detailed version of this algorithm, including explicit feature computation steps (performance metrics, perturbation magnitude, and optional sensitivity features), is provided in Appendix~\ref{app:algo_details}.} in Algorithm~\ref{algo:GP_membership_infer1}.   
\paragraph{Computation.}  
The method is efficient, requiring only forward evaluation of the target model (and optionally a backward pass if gradient features are used). The main cost comes from GP training, which scales as $\mathcal{O}(N^3)$ in the number of feature vectors $N$, though this is manageable for the modest sample sizes typical of membership inference.  

\begin{algorithm}[H]
\caption{GP-based membership inference}
\begin{algorithmic}[1]
\Require Trained model $f_{\theta^*}$, member set $\mathcal{D}_m$, non-member set $\mathcal{D}_n$
\Ensure Trained GP classifier $g$
\State Initialize empty feature set $\mathcal{M}$
\For{each $x \in \mathcal{D}_m \cup \mathcal{D}_n$}
    \State Compute diagnostic features $\phi(x)$ from $f_{\theta^*}$ (e.g.\ loss, entropy, margins, gradients, NTK)
    \State Label $y=1$ if $x \in \mathcal{D}_m$, else $y=0$
    \State Add $(\phi(x), y)$ to $\mathcal{M}$
\EndFor
\State Train GP classifier $g$ on $\mathcal{M}$
\State \Return $g$
\end{algorithmic}
\label{algo:GP_membership_infer1}
\end{algorithm}

\section{Experiments}

We evaluate the proposed GP-based membership inference (GP-MIA) method across three settings: a synthetic classification dataset, a real-world fraud detection task, an image classification task on CIFAR-10, and a language modeling task on WikiText-2. In each case, a supervised model is first trained on member data, diagnostic features are extracted from both member and non-member datasets, and a GP classifier is trained to distinguish membership status. For the synthetic, fraud detection and CIFAR-10 experiments, we use the common feature set (performance, confidence/entropy, perturbation magnitude, and input statistics). For the language model experiment, we additionally incorporate sensitivity features such as gradient norms and NTK-inspired quantities.

\subsection{Synthetic classification dataset}

\paragraph{Experimental setup}
We generated a binary classification dataset using \texttt{scikit-learn} \cite{SKlearn}. The member dataset contains 2000 balanced samples (50\%-50\%) from a 2-cluster Gaussian mixture with class separation 1.0 and 10 input features. The non-member dataset also contains 2000 samples but is more distinct, with separation 5.0, noise (label-flip probability 0.2), and class imbalance (80\%-20\%). A three-layer MLP was trained on the member dataset. After training, both member and non-member datasets were passed through the MLP to extract features (accuracy, entropy, perturbation magnitude, dataset statistics), which were used to train a GP classifier with variational inference.

We designed two experiments to probe how the GP classifier adapts. In the first, the GP was trained on the member dataset and the distinct non-member dataset (separation 5.0). It was then tested on three sets: (i) a 200-sample subset of the member dataset, (ii) a re-sampled dataset from the same member distribution (separation 1.0), and (iii) an intermediate dataset (separation 3.0, noise 0.2, 80-20 imbalance). In the second experiment, the non-member training set was augmented with additional samples closer to the member distribution (separation 3.0, balanced classes). The retrained GP was again tested on the same three sets.

\begin{figure}[ht]
  \centering
  \begin{minipage}[b]{0.44\textwidth}
    \centering
    \includegraphics[width=0.9\linewidth]{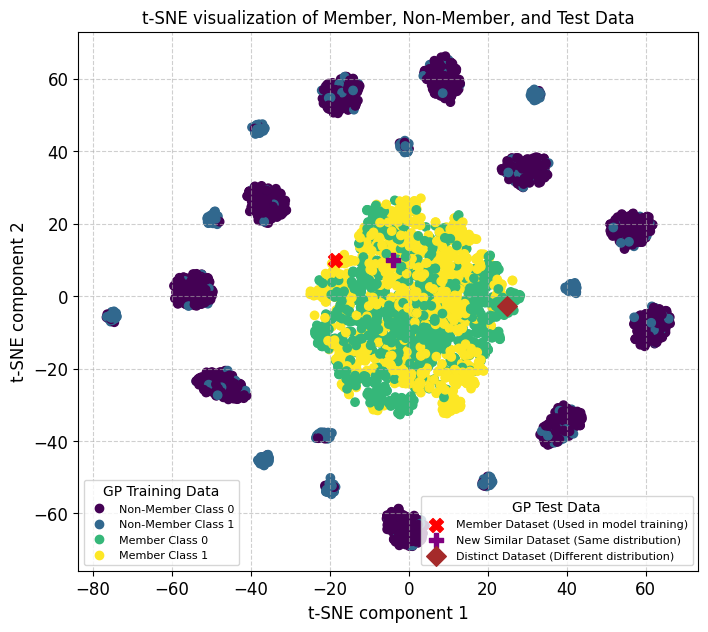}
    \caption*{(a) 2D visualization of feature space (t-SNE)}
  \end{minipage}
  \begin{minipage}[b]{0.55\textwidth}
    \centering
    \includegraphics[width=0.9\linewidth]{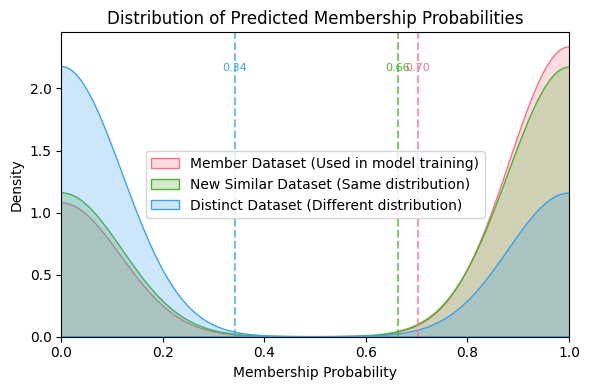}
    \caption*{(b) GP-predicted membership probability distributions}
  \end{minipage}
  \caption{Synthetic classification with large separation between member and non-member datasets.}
  \label{fig:synthetic1_1}
\end{figure}

\begin{figure}[ht]
  \centering
  \begin{minipage}[b]{0.44\textwidth}
    \centering
    \includegraphics[width=0.9\linewidth]{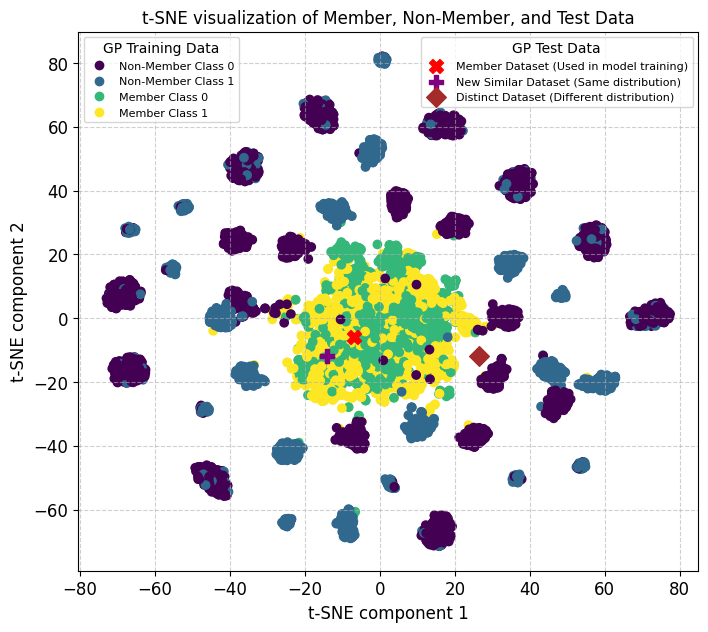}
    \caption*{(a) 2D visualization of feature space (t-SNE)}
  \end{minipage}
  \begin{minipage}[b]{0.55\textwidth}
    \centering
    \includegraphics[width=0.9\linewidth]{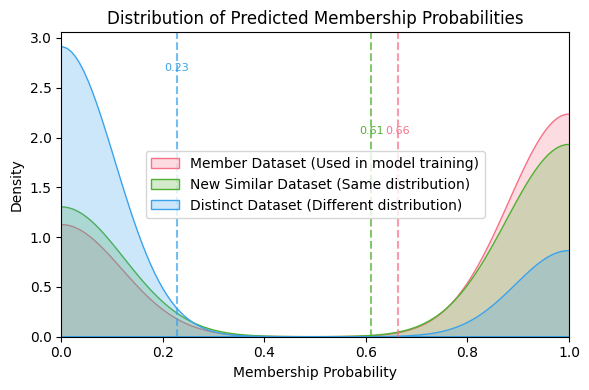}
    \caption*{(b) GP-predicted membership probability distributions}
  \end{minipage}
  \caption{Synthetic classification with smaller separation between member and non-member datasets.}
  \label{fig:synthetic1_2}
\end{figure}

\paragraph{Results}
In the first experiment (Figure~\ref{fig:synthetic1_1}b), the GP assigned high membership probabilities to the member and resampled datasets (means $\approx 0.7$), reflecting strong alignment with the training distribution. The intermediate dataset received lower probabilities (mean $\approx 0.34$) and greater variance, showing the model’s uncertainty for borderline cases. In the second experiment (Figure~\ref{fig:synthetic1_2}b), after training with augmented non-member data, the GP’s predictions shifted: member datasets still received high probabilities (means 0.61-0.66), but the intermediate dataset was pushed further away (mean $\approx 0.23$). This shows that informative non-member data improves the GP’s ability to differentiate ambiguous cases. Overall, the results highlight both the adaptability and calibrated uncertainty of the GP classifier.

\subsection{Real-world fraud detection}

\paragraph{Experimental setup}
We evaluate GP-MIA on the public \textit{Credit Card Fraud Detection} dataset from OpenML \cite{creditcard_fraud_openml}, a highly imbalanced tabular benchmark, which includes 284,807 transactions with only 492 positives. To construct a balanced working set, we use all fraud cases (positives) and down-sample an equal number of non-fraud cases (negatives), yielding 984 samples in total. This working set is stratified into training (60\%), validation (20\%), and test (20\%) splits. 
We train a single XGBoost classifier with balanced class weighting on the training split, using the validation split for monitoring. From these splits we construct dataset-level batches (each containing both fraud and non-fraud samples) labeled as either \textit{members} (drawn from the training split) or \textit{non-members} (drawn from the held-out test split). For each dataset $D$, we extract the \textit{common} post-hoc features with the frozen XGBoost model: accuracy, average prediction entropy, perturbation magnitude (average logit shift after a few additional boosting rounds on $D$), and dataset statistics (mean absolute feature means and mean feature variances). Feature vectors are standardized and used to train a Gaussian Process classifier with an RBF $+$ white noise kernel. We train the GP on 80\% of the dataset-level examples and evaluate on the remaining 20\%.  

\begin{figure}[ht]
  \centering
  \begin{minipage}[b]{0.44\textwidth}
    \centering
    \includegraphics[width=0.9\linewidth]{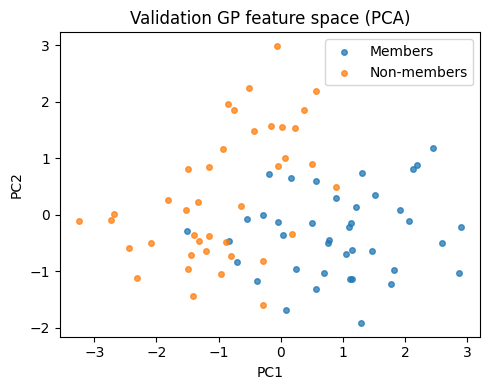}
    \caption*{(a) PCA of GP features (validation)}
  \end{minipage}
  \hfill
  \begin{minipage}[b]{0.55\textwidth}
    \centering
    \includegraphics[width=0.9\linewidth]{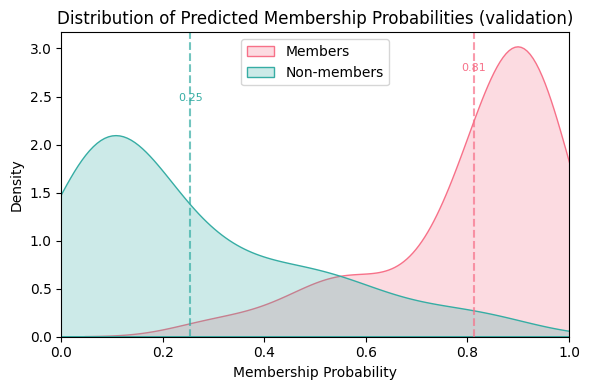}
    \caption*{(b) GP-predicted membership probability distributions}
  \end{minipage}
  \caption{Credit card fraud detection with GP-MIA. Members and non-members separate clearly in both feature space (PCA) and predicted probability distributions.}
  \label{fig:fraud_results}
\end{figure}

\paragraph{Results}
The GP classifier achieves strong discrimination: AUROC $=0.959$, AUPR $=0.961$, and TPR@1\%FPR $=0.60$. Confusion matrix at the default $0.5$ threshold confirms balanced precision ($0.84$-$0.92$) and recall ($0.83$-$0.93$) across member and non-member groups. Table~\ref{tab:fraud_results} summarizes the validation performance. Figure~\ref{fig:fraud_results}b shows the predicted membership probability distributions: members concentrate near higher probabilities (mean $\approx 0.81$), while non-members cluster around lower values (mean $\approx 0.25$), demonstrating clear separation. A PCA visualization of the GP feature space (Figure~\ref{fig:fraud_results}a) shows that members and non-members form distinguishable clusters in two dimensions, supporting the separability of the extracted features. 
These results indicate that even on a challenging, real-world financial dataset with limited positives and high class imbalance, GP-MIA successfully distinguishes member from non-member datasets using only post-hoc metrics, without requiring shadow models or internal access.

\begin{table}[ht]
\centering
\caption{Performance of GP-MIA on credit card fraud detection (validation set).}
\label{tab:fraud_results}
\begin{tabular}{lccc}
\toprule
Metric & Value \\
\midrule
AUROC & 0.959 \\
AUPR & 0.961 \\
TPR @ 1\% FPR & 0.600 \\
\midrule
\multicolumn{2}{l}{Confusion Matrix (threshold 0.5)} \\
\midrule
 & Predicted Non-member & Predicted Member \\
True Non-member & 33 & 7 \\
True Member     & 3  & 37 \\
\bottomrule
\end{tabular}
\end{table}

\subsection{CIFAR-10 classification}

\paragraph{Experimental setup}
We next evaluated GP-MIA on the CIFAR-10 dataset. Ten member datasets were created by sampling 500 non-overlapping images each from the CIFAR-10 training set. For non-members, we considered two categories: (i) ten distinct datasets from the SVHN dataset, and (ii) ten similar datasets created by applying strong Gaussian noise (strength 0.4) to unseen CIFAR-10 images. A convolutional neural network (CNN) was trained for 20 epochs on all member datasets combined. From each dataset, we extracted common features: classification accuracy, average prediction entropy, average loss, prediction confidence, correct-class confidence, perturbation magnitude (via 5-epoch fine-tuning), and input statistics (mean, standard deviation, variance of pixels). These were used to train a GP classifier with an RBF + white-noise kernel and variational inference.

We tested the GP on five scenarios: (i) a member dataset from training, (ii) a new CIFAR-10 dataset from the same distribution, (iii) an SVHN dataset, (iv) an augmented CIFAR-10 dataset, and (v) an interpolated dataset mixing CIFAR-10 and SVHN (60-40).

\begin{figure}[ht]
  \centering
  \begin{minipage}[b]{0.44\textwidth}
    \centering
    \includegraphics[width=0.9\linewidth]{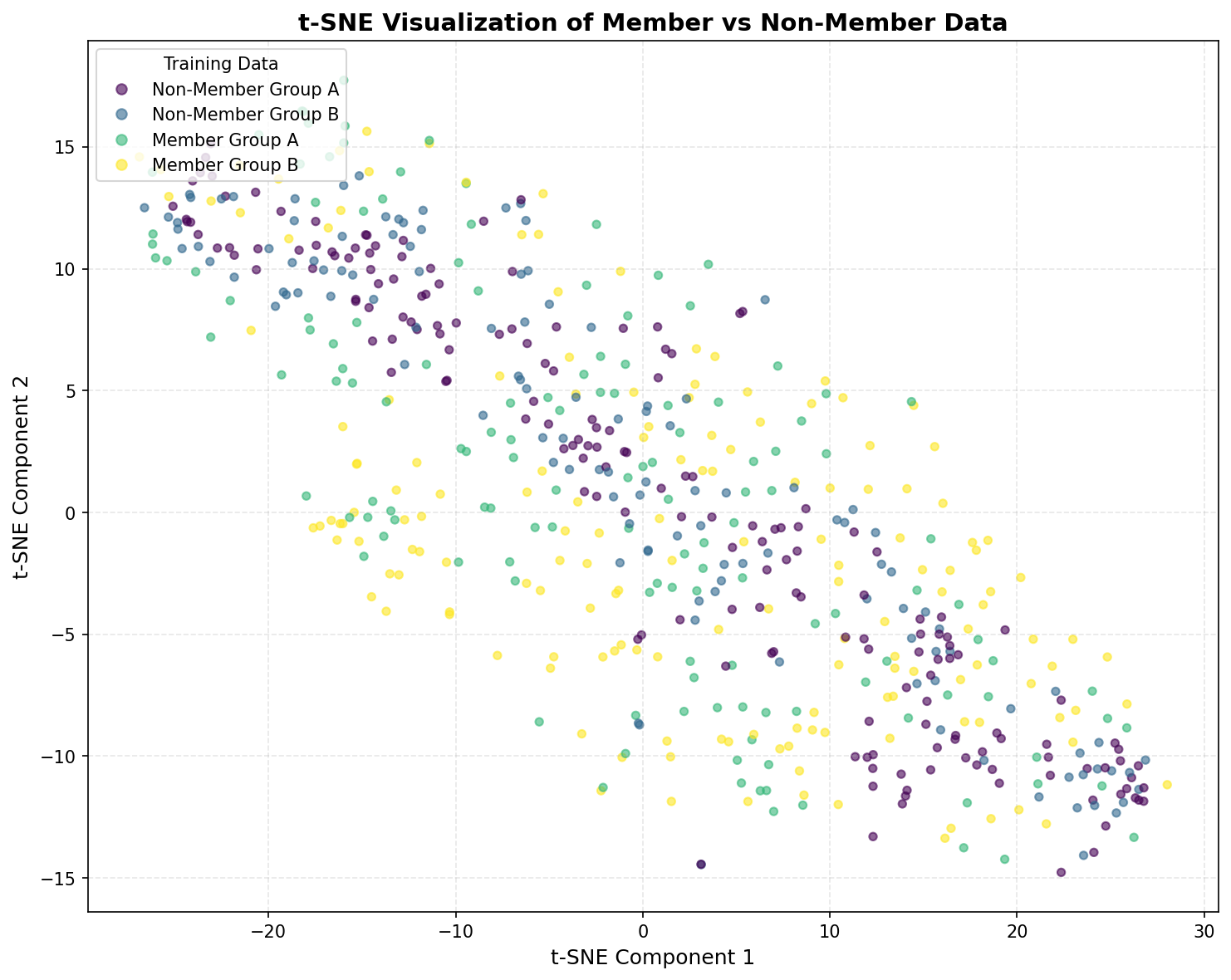}
    \caption*{(a) 2D visualization of feature space (t-SNE)}
  \end{minipage}
  \begin{minipage}[b]{0.55\textwidth}
    \centering
    \includegraphics[width=0.9\linewidth]{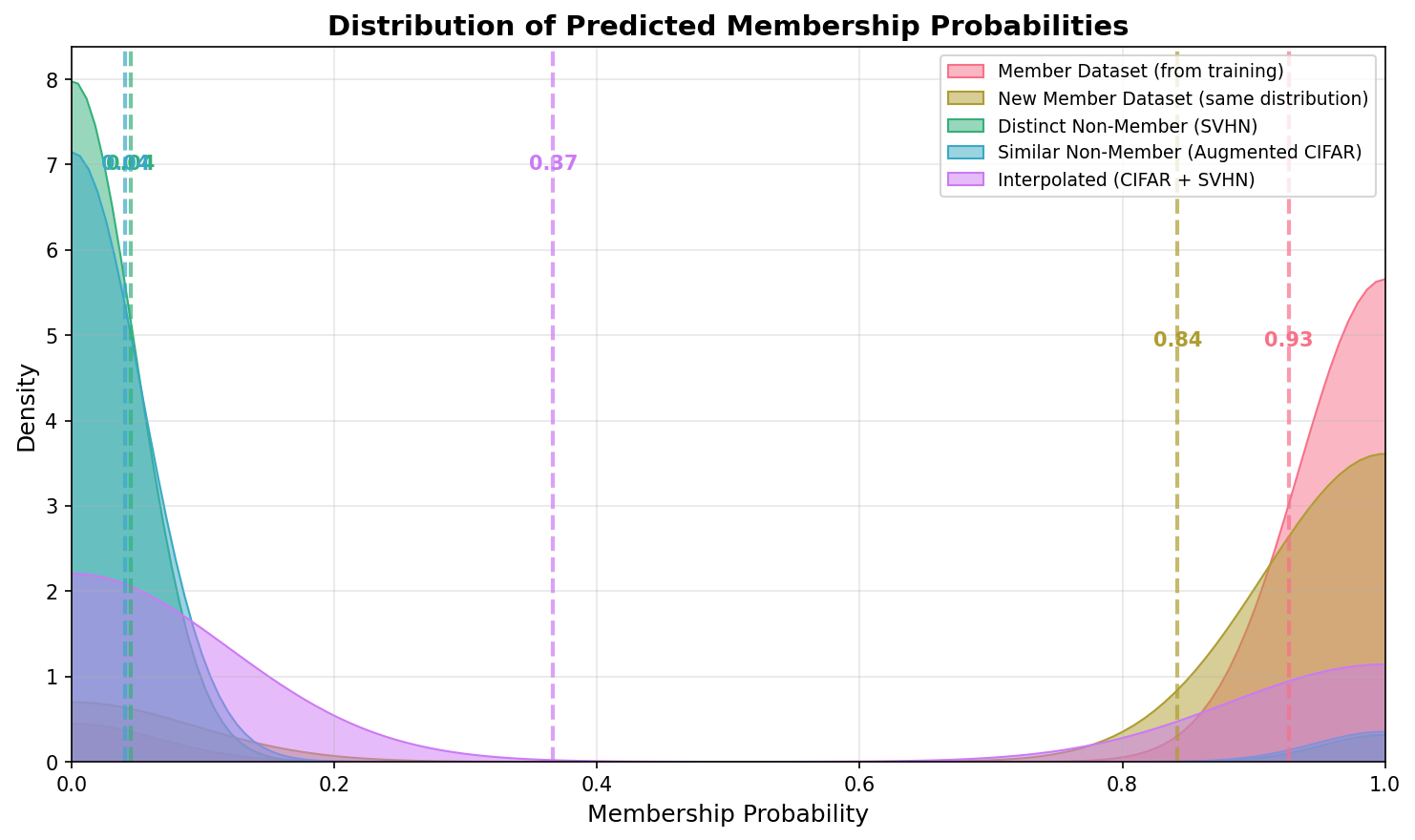}
    \caption*{(b) GP-predicted membership probability distributions}
  \end{minipage}
  \caption{Membership inference on CIFAR-10. GP classifier distinguishes member datasets from both similar and distinct non-members, while expressing uncertainty for interpolated cases.}
  \label{fig:cifar}
\end{figure}

\paragraph{Results}
Figure~\ref{fig:cifar}b shows the distribution of GP predictions. The training member dataset received the highest membership probability (0.93), while a new CIFAR-10 set scored 0.84, confirming generalization to the underlying distribution. Both distinct (SVHN) and similar (augmented CIFAR-10) non-member datasets were confidently classified as non-members ($\approx 0.04$). The interpolated dataset received an intermediate probability (0.37), reflecting appropriate uncertainty at the decision boundary. The t-SNE visualization (Figure~\ref{fig:cifar}a) shows overlap between member and non-member feature distributions, underlining why probabilistic classification is needed beyond simple geometric separation. These results demonstrate that GP-MIA can effectively detect training set membership and quantify uncertainty across varying distributional shifts.

\subsection{Language modeling on WikiText-2}

\paragraph{Experimental setup}
Finally, we evaluated GP-MIA on a transformer-based language model trained on WikiText-2 for next-token prediction. The corpus was tokenized using the GPT-2 tokenizer, with sequence length 128. We trained a compact GPT-2 style model (3 layers, 4 heads, embedding dimension 192) for 300 steps using AdamW with cosine scheduling. Member sequences were drawn from the training set, and non-member sequences from the held-out test set, each with 800 examples.

For each sequence, we extracted both common and sensitivity features: negative log-likelihood, prediction entropy, logit margin (difference between true and top competing token), perplexity, and gradient norm of the final normalization layer. Features were normalized and used to train a GP classifier with an RBF + white-noise kernel.

\begin{figure}[ht]
  \centering
  \begin{minipage}[b]{0.44\textwidth}
    \centering
    \includegraphics[width=0.9\linewidth]{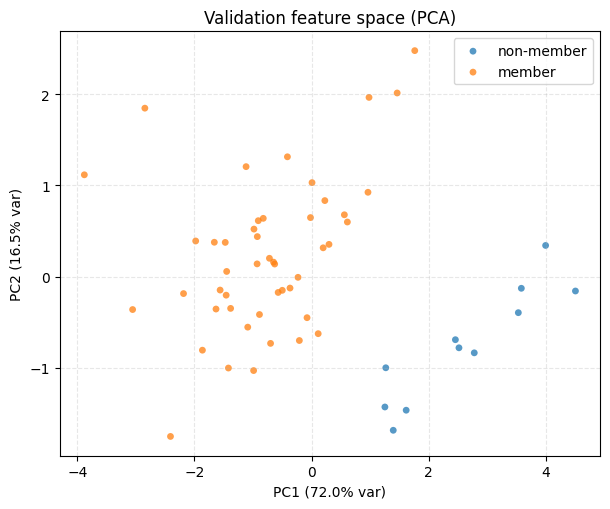}
    \caption*{(a) 2D visualization of feature space (PCA)}
  \end{minipage}
  \begin{minipage}[b]{0.55\textwidth}
    \centering
    \includegraphics[width=0.9\linewidth]{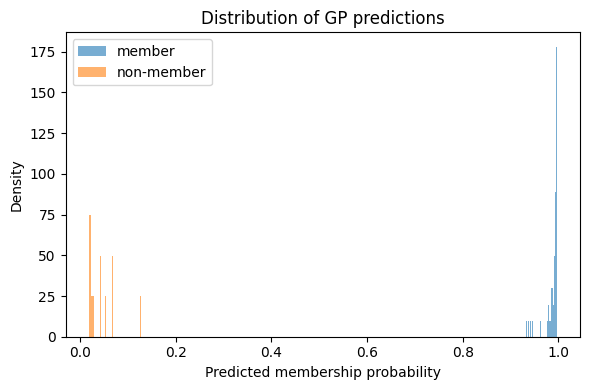}
    \caption*{(b) GP-predicted membership probability distributions}
  \end{minipage}
  \caption{Membership inference on WikiText-2 language model. Members and non-members form well-separated clusters in feature space, aligning with bimodal GP membership predictions.}
  \label{fig:lm_results}
\end{figure}

\paragraph{Results}
The GP classifier achieved near-perfect performance (AUROC = 1.000, AUPR = 1.000, TPR@1\%FPR = 1.000). Confusion matrices showed zero misclassifications across thresholds. PCA visualization (Figure~\ref{fig:lm_results}a) revealed strong separation between member and non-member features, consistent with the sharply bimodal GP membership probability distributions (Figure~\ref{fig:lm_results}b). Table~\ref{tab:lm_mia_results} summarizes the results. These findings confirm that even small language models leak strong membership signals detectable by GP-MIA, particularly when sensitivity features are included.

\begin{table}[ht]
\centering
\caption{Validation performance of GP classifier on WikiText-2 membership inference.}
\label{tab:lm_mia_results}
\begin{tabular}{lccc}
\toprule
 & AUROC & AUPR & TPR@1\%FPR \\
\midrule
GP Classifier & 1.000 & 1.000 & 1.000 \\
\bottomrule
\end{tabular}
\end{table}

\section{Discussion}

Our findings show that Gaussian Process meta-modeling provides a flexible and data-efficient framework for membership inference. Operating in a post-hoc manner, GP-MIA avoids the overhead of shadow models and repeated queries while still capturing informative distributional signals. Lightweight features such as entropy and dataset statistics already offer strong discrimination, while sensitivity-based features (e.g., gradients, NTK scores) further improve separability in complex models such as language models.  

Beyond privacy attacks, GP-MIA can be repurposed for detecting distributional shifts, monitoring generalization, or auditing deployed models. Scaling to larger datasets and high-dimensional models will require efficient kernel approximations, and exploring richer feature spaces may further enhance separability. From a defense perspective, minimizing detectable differences in post-hoc metrics, or adopting architectures that re-encode inputs, could help mitigate membership leakage. More broadly, this work highlights the utility of Gaussian processes as both a privacy attack and a diagnostic tool for trustworthy machine learning.  

\section{Conclusion}

We presented GP-MIA, a Gaussian process-based approach to membership inference that leverages post-hoc metrics from a single trained model. 
By combining lightweight features such as accuracy, entropy, and dataset statistics with optional sensitivity-based measures, GP-MIA achieves efficient and interpretable membership predictions without requiring shadow models, retraining, or extensive query access. 
A key strength of the GP framework is its ability to provide calibrated uncertainty estimates, which quantify the confidence of each prediction and improve interpretability. 
The method is efficient since it operates entirely in a post-hoc manner. 
Experiments on synthetic benchmarks, real-world financial fraud detection, CIFAR-10, and WikiText-2 confirm that GP-MIA delivers strong and consistent performance across diverse domains, establishing it as a practical and versatile tool for auditing privacy risks in machine learning models.

\paragraph{Outlook} Future work includes exploring alternative kernel choices, developing scalable approximations for large models and datasets, and integrating GP-MIA into broader privacy defense frameworks to inform both attacks and mitigations.

\bibliographystyle{plain}
\bibliography{reference}

\newpage

\appendix

\section{Detailed GP-MIA algorithm}
\label{app:algo_details}

For completeness, we provide here a detailed version of the GP-MIA procedure, expanding Algorithm.\ref{algo:GP_membership_infer1} with explicit feature construction steps.

\begin{algorithm}[H]
\caption{GP-based membership inference (GP-MIA)}
\begin{algorithmic}[1]
\Require Trained supervised model $f$, member datasets $\mathcal{D}_{\text{member}}$, non-member datasets $\mathcal{D}_{\text{non-member}}$
\Ensure Trained GP classifier $g$
\State Initialize empty feature-label set $\mathcal{M}$
\For{each dataset or sample $D$ in $\mathcal{D}_{\text{member}} \cup \mathcal{D}_{\text{non-member}}$}
    \State Pass $D$ through $f$ to obtain predictions $(\hat{y}, p(y|x))$
    \State Compute \textit{common features}:
        \begin{itemize}
            \item Performance metrics (e.g. accuracy for classification, MSE for regression)
            \item Model confidence (e.g. average prediction entropy, negative log-likelihood, perplexity)
            \item Input statistics (mean and variance of $x$)
        \end{itemize}
    \State Compute \textit{perturbation magnitude}:
        \begin{itemize}
            \item Create a copy of $f$ and lightly fine-tune on $D$ for a few epochs
            \item Measure $\ell_2$ distance between original and fine-tuned parameters
        \end{itemize}
    \State Optionally compute \textit{sensitivity features} (for neural networks):
        \begin{itemize}
            \item Gradient norms: $\|g_\theta(x)\|_{\mathrm{F}}, \|g_\ell(x,y)\|_2$
            \item Input-Jacobian norm: $\|J_x(x)\|_{\mathrm{F}}$
            \item NTK statistics: ridge leverage $\tau_\lambda(x)$, projection scores $h_\lambda(x)$, similarity measures $s_{\max}(x), \bar{s}(x)$
        \end{itemize}
    \State Concatenate all metrics into feature vector $\mathbf{m}_D$
    \State Label $\mathbf{m}_D$ as member (1) or non-member (0)
    \State Add $(\mathbf{m}_D, \text{label})$ to $\mathcal{M}$
\EndFor
\State Train GP classifier $g$ with RBF + white-noise kernel on $\mathcal{M}$
\State \Return $g$
\end{algorithmic}
\label{algo:GP_membership_infer2}
\end{algorithm}

\end{document}